\documentclass{ifacconf}

\usepackage{graphicx}      
\usepackage{natbib}        

\usepackage{amsmath}
\usepackage{amssymb}
\usepackage{subcaption}
\usepackage{booktabs}
\usepackage{bm}
\usepackage{siunitx}

\newcommand{\R}{{\mathbb{R}}}

\newcommand{\Rn}{{\mathbb{R}}^{n}}
\newcommand{\Rm}{{\mathbb{R}}^{m}}
\newcommand{\Rd}{{\mathbb{R}}^{d}}

\newcommand{\RRm}{{\mathbb{R}^{2m}}}

\newcommand{\Rnn}{{\mathbb{R}^{n \times n}}}

\newcommand{\T}{{\text{T}}}     

\newcommand{\pdot}{{\Dot{p}}}
\newcommand{\qdot}{{\Dot{q}}}

\newcommand{\xdot}{{\Dot{x}}}

\newcommand{\eye}{{\bm{I}}}
\newcommand{\zero}{{\bm{0}}}

\newcommand{\inv}[1]{{#1}^{-1}}

\newcommand{\RKHS}{\mathcal{H}_{K}}
\newcommand{\innerproduct}[2]{\left\langle #1 , #2 \right\rangle}

\newcommand{\beq}{\begin{equation}}
\newcommand{\eeq}{\end{equation}}
\newcommand{\bb}{\begin{bmatrix}}
\newcommand{\eb}{\end{bmatrix}}

\DeclareMathOperator*{\argmin}{\arg\min}
\DeclareMathOperator{\spn}{span}

\newcommand{\boldfhat}{\hat{\bm{f}}}

\newcommand{\boldpdot}{\Dot{\boldsymbol{p}}}
\newcommand{\boldqdot}{\Dot{\boldsymbol{q}}}

\newcommand{\boldxdot}{\Dot{\boldsymbol{x}}}

\newcommand{\boldalpha}{\bm{\alpha}}
\newcommand{\boldbeta}{\bm{\beta}}

\newcommand{\boldnabla}{\bm{\nabla}}
\newcommand{\boldxi}{\bm{\xi}}

\newcommand{\boldtau}{\bm{\tau}}

\newcommand{\boldphi}{\bm{\phi}}

\newcommand{\boldLambda}{\bm{\Lambda}}

\newcommand{\boldPhi}{\bm{\Phi}}
\newcommand{\boldPsi}{\bm{\Psi}}

\newcommand{\bolda}{{\bm{a}}}
\newcommand{\boldb}{{\bm{b}}}
\newcommand{\boldc}{{\bm{c}}}

\newcommand{\boldf}{{\bm{f}}}
\newcommand{\boldg}{{\bm{g}}}

\newcommand{\boldp}{{\bm{p}}}
\newcommand{\boldq}{{\bm{q}}}

\newcommand{\boldw}{{\bm{w}}}
\newcommand{\boldx}{{\bm{x}}}
\newcommand{\boldy}{{\bm{y}}}
\newcommand{\boldz}{{\bm{z}}}

\newcommand{\boldG}{{\bm{G}}}

\newcommand{\boldJ}{{\bm{J}}}
\newcommand{\boldK}{{\bm{K}}}

\newcommand{\boldR}{{\bm{R}}}

\newcommand{\boldX}{{\bm{X}}}

\newcommand{\calH}{{\mathcal{H}}}

\newcommand{\calN}{{\mathcal{N}}}

\newcommand{\calZ}{{\mathcal{Z}}}

\makeatletter
\newcommand\thefontsize{\textit{The current font size is: \f@size pt}}
\makeatother

\begin{document}


\begin{frontmatter}

\title{Learning dissipative Hamiltonian dynamics with reproducing kernel Hilbert spaces and random Fourier features}


\author[First]{Torbjørn Smith}
\author[First]{Olav Egeland}

\address[First]{Department of Mechanical and Industrial Engineering, Norwegian University of Science and Technology (NTNU), NO-7491 Trondheim, Norway. (email: torbjorn.smith@ntnu.no; olav.egeland@ntnu.no)}

\begin{abstract}                
    This paper presents a new method for learning dissipative Hamiltonian dynamics from a limited and noisy dataset. The method uses the Helmholtz decomposition to learn a vector field as the sum of a symplectic and a dissipative vector field. The two vector fields are learned using two reproducing kernel Hilbert spaces, defined by a symplectic and a curl-free kernel, where the kernels are specialized to enforce odd symmetry. Random Fourier features are used to approximate the kernels to reduce the dimension of the optimization problem. The performance of the method is validated in simulations for two dissipative Hamiltonian systems, and it is shown that the method improves predictive accuracy significantly compared to a method where a Gaussian separable kernel is used.
\end{abstract}


\end{frontmatter}


\section{INTRODUCTION}\label{sec:1_introduction}
\vspace{-2.5mm}
Data-driven modeling of dynamical systems is a fundamental task in robotics and control engineering, and machine learning is emerging as a powerful alternative to deriving models analytically from first principles \citep{Brunton2022}. Even though data-driven methods are powerful and expressive, it is still a challenge to produce models that generalize well and satisfy physical laws. Overfitting may occur if the data used to learn the models is limited and corrupted by noise. Physics-informed machine learning can be used where structure and constraints are enforced on the learning problem to remedy such challenges. Energy conservation and Hamiltonian dynamics are physical laws that can be enforced in the learning of dynamical systems using the Hamiltonian formalism \citep{Greydanus2019}. This is effective for producing accurate and generalizable models, and of high interest from a control perspective \citep{Ahmadi2018}. It is interesting to extend these results to systems with dissipative terms, which was done with a Helmholtz decomposition in \citep{Sosanya2022}.

\textit{Related work:} \citet{Ahmadi2018} performed control-oriented learning of Lagrangian and Hamiltonian dynamics from limited trajectory data. They used polynomial basis functions and solved the learning problem using quadratic programming. By learning the Lagrangian and Hamiltonian functions, they could accurately learn the dynamics from a limited number of trajectories. \citet{Greydanus2019} proposed a method for learning the Hamiltonian dynamics of energy-conserving systems. The dynamics were learned by taking the symplectic gradient of a learned Hamiltonian function parameterized by a neural net. This significantly improved the predictive accuracy of the learned model. \citet{Zhong2020} expanded on the work by \citet{Greydanus2019} by eliminating the need for higher-order derivatives in the data set and including the option for energy-based control. \citet{Chen2020} improved on the work by \citet{Greydanus2019} by integrating the partial derivatives of the learned Hamiltonian by using a symplectic integrator and by back-propagating over multiple time steps. This improved the learning of more complex and noisy Hamiltonian systems.

\citet{Zhong2020dissipative} expanded on the work presented in \citep{Zhong2020} to include energy dissipation to learn dissipative Hamiltonian dynamics, which improved prediction accuracy over a naive approach. \citet{Sosanya2022} expanded on the work in \citep{Greydanus2019} by using the Helmholtz decomposition to learn dissipative Hamiltonian dynamics. They utilized two neural nets to learn two separate scalar-valued functions, one for the energy-conserving Hamiltonian mechanics and one for the dissipative gradient dynamics, resulting in an additive final learned dynamical model. The approach improved prediction accuracy and allowed for the investigation of varying levels of energy dissipation.

\citet{Smith2024} performed system identification on Hamiltonian mechanical systems using an odd symplectic kernel to enforce energy conservation and odd symmetry. The vector fields were learned over a reproducing kernel Hilbert space defined by the odd symplectic kernel and approximated using random Fourier features for dimensionality reduction, faster training, and inference speed. The method showed that the side constraints enforced through the kernel improved prediction accuracy, generalizability, and data efficiency.

\textit{Contribution:} In this paper, we show how dissipative Hamiltonian dynamical systems can be learned using functions in a reproducing kernel Hilbert space with random Fourier features. The dynamical systems are learned using a Helmholtz decomposition to produce an additive model consisting of a symplectic and a dissipative part. The symplectic part is learned using a symplectic kernel, and the dissipative part is learned using a curl-free kernel. To further aid the generalized performance of the learned model, both kernels are modified to enforce odd symmetry in the domain of the learned function. By learning over a reproducing kernel Hilbert space, a general and expressive approach is presented, and due to the random Fourier features and the structure of the method, computational efficiency, accuracy, and generalizability are achieved. The method is validated in two simulation examples where the dissipative Hamiltonian dynamics of two mechanical systems are learned and compared to a baseline implementation.
\section{PRELIMINARIES}\label{sec:2_preliminaries}
\vspace{-2mm}
The aim of this paper is to learn the unknown dynamic system given by 
\begin{equation}\label{vector_field_dynamic_model}
    \boldxdot = \boldf(\boldx)
\end{equation}
 where ${\boldx \in \Rn}$ is the state vector, ${\boldxdot \in \Rn}$ is the time derivative of the state vector, and ${\boldf : \Rn \rightarrow \Rn}$ are the system dynamics. Given a set of ${N}$ data points ${ \{  (\boldx_{i}, \boldxdot_{i})  \in \Rn \times \Rn \}_{i=1}^{N}}$ the aim is to learn a function ${\hat\boldf \in \RKHS}$, where the class of functions ${\RKHS}$ is a reproducing kernel Hilbert space (RKHS).

\subsection{Reproducing kernel Hilbert space}

The theory on reproducing kernels was presented by \citet{Aronszajn1950}, and the extension to vector-valued functions was made in \citep{Micchelli2005} and \citep{Minh2011}. In this paper, a vector-valued function ${\boldf : \Rn \rightarrow \Rn}$ is to be learned. Let $\calH_K$ be the RKHS defined by the matrix-valued reproducing kernel ${\boldK : \Rn \times \Rn \rightarrow \Rnn}$. The reproducing kernel $\boldK$ is required to be positive definite, which means that $\boldK(\boldx,\boldz) = \boldK(\boldz,\boldx)^\T$ for all $\boldx, \boldz \in \Rn$, and $\sum_{i=1}^{N} \sum_{j=1}^{N} \innerproduct{\bolda_{i}}{\boldK(\boldx_{i},\boldx_{j})\bolda_{j}} \geq 0$ for any set of vectors ${\{ \boldx_i \}_{i=1}^{N}, \{\bolda_i \}_{i=1}^{N} \in \Rn}$. Then ${\boldK}$ is a reproducing kernel and defines an RKHS ${\RKHS}$. This corresponds to the Moore-Aronszajn theorem \citep{Aronszajn1950} in the scalar case. Define the function ${\boldK_{x}\bolda : \Rn \rightarrow \Rn}$ by
\begin{equation}
    (\boldK_{x}\bolda)(\boldz) = \boldK(\boldz,\boldx)\bolda \in \Rn, \quad \forall \; \boldz \in \Rn
\end{equation}
The notation $\boldK_{x} = \boldK(\cdot,\boldx)$ will also be used. The ${\RKHS}$ is the closure of 
\begin{equation}
   \spn\{ \boldK_{x}\bolda \; | \; \boldx \in \Rn, \bolda \in \Rn\}
    \subseteq \RKHS
\end{equation}
Then, functions in the RKHS $\RKHS$ can be defined as ${\boldf = \sum_{i=1}^{N} \boldK_{x_i}\bolda_i \in \RKHS}$ and ${\boldg = \sum_{j=1}^{N} \boldK_{z_j}\boldb_j \in \RKHS}$ with inner product
\begin{equation}\label{eq:inner_prod_func_in_rkhs}
    \innerproduct{\boldf}{\boldg}_{\RKHS} = \sum_{i=1}^{N} \sum_{j=1}^{N} \innerproduct{\bolda_{i}}{\boldK(\boldx_{i},\boldz_{j})\boldb_{j}}
\end{equation}
and norm $\|\boldf\|_{\RKHS}^2 = \innerproduct{\boldf}{\boldf}_{\RKHS}$. The reproducing property is given by
\begin{equation}\label{eq:reproducing_property_vector}
    \innerproduct{\boldf}{ \boldK_{x}\bolda}_{\RKHS}
    = \innerproduct{\sum_{i=1}^{N} \boldK_{x_i}\bolda_i} {\boldK_x\bolda}
    = \innerproduct{\boldf(\boldx)}{\boldy}
\end{equation}  
where it is used that $\boldf(\boldx) = \sum_{i=1}^{N} \boldK(\boldx,\boldx_i)\bolda_i$. It is noted that if ${\| \boldf -\boldg\|_{\RKHS}}$ converges to zero, then ${\|\boldf(\boldx) - \boldg(\boldx)}\|$ converges to zero for each ${\boldx}$.

\subsection{Learning dynamical systems with RKHS}

The unknown vector field is estimated using the vector-valued regularized least-squares problem \citep{Micchelli2005}
\begin{equation}\label{eq:vector_valued_regular_least_squares}
    \hat\boldf = \argmin_{\boldf \in \RKHS} \frac{1}{N} \sum_{i=1}^{N} \| \boldf(\boldx_{i}) - \boldxdot_{i} \|^{2} + \lambda \| \boldf \|_{\RKHS}^{2}
\end{equation}
where ${\calZ = \{ (\boldx_{i},\boldxdot_{i}) \in \Rn \times \Rn \}_{i=1}^{N}}$ is the data used to learn the vector field, and ${\lambda > 0}$ is the regularization parameter. By the representer theorem \citep{Scholkopf2001}, the function is given by ${\boldfhat = \sum_{i=1}^{N} \boldK(\cdot,\boldx_i) \bolda_i \in \RKHS}$ where the optimal solution is given by the coefficients ${\{ \bolda_i \}_{i=1}^{N} \in \Rn}$ found from \citep{Micchelli2005}
\begin{equation}\label{eq:vector_valued_regular_least_squares_a_equation}
    \sum_{j=1}^N \boldK(\boldx_i,\boldx_j)\bolda_j + N\lambda \bolda_i = \boldxdot_i
\end{equation}
and the function value is 
\begin{equation}\label{eq:vector_valued_regular_least_squares_f_equation}
    \hat\boldf(\boldx) = \sum_{i=1}^{N} \boldK(\boldx,\boldx_i) \boldsymbol{a}_i \in \Rn
\end{equation}
A matrix formulation of \eqref{eq:vector_valued_regular_least_squares_a_equation} is found in \citep{Minh2011}.

\subsection{Random Fourier features}

The dimension of the learning problem in \eqref{eq:vector_valued_regular_least_squares} increases with the number of samples in the training set. This increases the training time for the model and the inference time of the learned model in \eqref{eq:vector_valued_regular_least_squares_f_equation}. To limit training and inference time, random Fourier features (RFF) are used to approximate the kernel functions associated with functions in an RKHS \citep{Singh2021}.

A matrix-valued feature map ${\boldPsi : \Rn \rightarrow \R^{d \times n}}$ is used to approximate the matrix-valued kernel function ${\boldK}$ \citep{Sindhwani2018} as ${\boldK(\boldx,\boldy) \approx \boldPsi(\boldx)^{\T} \boldPsi(\boldy)}$. The function value of the vector field in \eqref{eq:vector_valued_regular_least_squares_f_equation} can then be parameterized as
\begin{equation}\label{eq:rkhf_f_using_rff}
    \boldfhat(\boldx) \approx \sum_{i=1}^{N} \boldPsi(\boldx)^{\T} \boldPsi(\boldx_i) \bolda_i = \boldPsi(\boldx)^{\T} \boldalpha
\end{equation}
where the new model coefficients are given as $\boldalpha = \sum_{i=1}^{N} \boldPsi(\boldx_i) \bolda_i \in \Rd$. The number of random features ${d}$ is chosen by balancing the quality of the approximation and the computational time of the function. Using \eqref{eq:inner_prod_func_in_rkhs} and \eqref{eq:rkhf_f_using_rff} the norm of ${\boldf \in \RKHS}$ can be written as ${\| \boldf \|_{\mathcal{H}_{K}}^{2} = \| \boldalpha \|^{2}}$. The optimization problem in \eqref{eq:vector_valued_regular_least_squares} is then re-formulated as an optimization problem over ${\boldalpha \in \Rd}$
\begin{equation}\label{eq:rff_vector_valued_regular_least_squares}
    \boldalpha^{*} = 
    \argmin_{\boldalpha \in \Rd} \frac{1}{N} \sum_{i=1}^{N} \| \boldPsi(\boldx_i)^{\T} \boldalpha - \boldxdot_{i} \|^{2} + \lambda \boldalpha^{\T} \boldalpha
\end{equation}
and the function value of the optimal vector field is given by \eqref{eq:rkhf_f_using_rff}.

\subsection{Hamiltonian dynamics}

Consider a holonomic system with generalized coordinates $\boldq \in \Rm$, generalized momenta $\boldp \in \Rm$, and Hamiltonian $H(\boldq,\boldp)$ \citep{Goldstein2002}. Hamilton's equations of

motion are given by
\begin{align}
    \boldqdot &= \boldnabla_p H \label{eq:hamilton_gen_coord_dot}\\
    \boldpdot &= -\boldnabla_q H + \boldtau \label{eq:hamilton_gen_momenta_dot}
\end{align}
where $\boldtau \in \Rm$ are the input generalized forces. The operators $\boldnabla_p = \partial/\partial \boldp$ and $\boldnabla_q = \partial/\partial \boldq$ are column vectors.

This can be formulated in the phase space with state vector ${\boldx = \bb \boldq^{\T},\boldp^{\T} \eb^{\T} \in \RRm}$. Hamilton's equations of motion \eqref{eq:hamilton_gen_coord_dot}, \eqref{eq:hamilton_gen_momenta_dot} with $\boldtau=\zero$ are then
\begin{equation}\label{eq:Hamiltonian_dynamics_J}
    \boldxdot = \boldf(\boldx) = \boldJ \boldnabla  H (\boldx)
\end{equation}
where $\boldnabla = [\boldnabla_q^\T,\boldnabla_p^\T]^\T = \partial/\partial \boldx$, ${\boldf : \RRm \rightarrow \RRm}$ is a vector field and 
\beq\label{eq:symplectic_matrix}
    \boldJ = 
    \bb 
        \;\zero\; & \eye_m\\
        -\eye_m \;  & \zero
    \eb
    \in \R^{2m \times 2m}
\eeq
is the symplectic matrix. The divergence of the vector field $\boldf$ is zero, which is seen from
\begin{equation}
    \boldnabla^\T \boldf(\boldx) = \boldnabla^\T \boldJ  \boldnabla  H (\boldx) = 0 
\end{equation}
since $\boldnabla^\T\boldJ \boldnabla = 0$. This implies that the volume of the phase flow is preserved \citep{Arnold1989}.

The flow of \eqref{eq:Hamiltonian_dynamics_J} in the phase space with initial condition $\boldx(0) = \boldx_0$ is given by $\boldphi_{t}(\boldx_0)$. Define the Jacobian ${\boldPsi(t) = \partial \boldphi_{t}(\boldx_0) / \partial \boldx_{0}}$. Then the system \eqref{eq:Hamiltonian_dynamics_J} is said to be symplectic if 
\beq\label{eq:symplectic_condition}
    \boldPsi(t)^{\T} \boldJ \boldPsi(t) = \boldJ
\eeq
for all $t\geq 0$. A system is Hamiltonian if and only if it is symplectic \citep{Hairer2006}. 

\subsection{Dissipative Hamiltonian dynamics}

In the case where there is dissipation in addition to the Hamiltonian dynamics, the dynamics can be written as
\begin{equation}\label{eq:Hamiltonian_dynamics_J_R}
    \boldxdot = \boldf(\boldx) = (\boldJ - \boldR) \boldnabla  H (\boldx)
\end{equation}
where $\boldR$ is symmetric and positive semidefinite. This formulation is used in port-Hamiltonian systems \citep{Ortega2008}. The divergence of the vector field is then  
\begin{equation}
   \boldnabla^\T \boldf(\boldx) 
= - \boldnabla^\T \boldR  \boldnabla  H (\boldx) 
\end{equation}
where $\boldnabla^\T \boldR  \boldnabla$ is not necessarily zero. The time derivative of the Hamiltonian is 
\begin{equation}
    \dot H = - (\boldnabla H)^\T \boldR \boldnabla H  \leq 0
\end{equation}
If $H$ is the energy of the system, then the energy dissipation of the system is due to the vector field $\boldR\boldnabla H$.

The vector field in \eqref{eq:Hamiltonian_dynamics_J_R} can be written as
\beq\label{eq:Helmholtz decomposition}
\boldf(\boldx) = \boldf_s(\boldx) + \boldf_d (\boldx)
\eeq
where $\boldf_s(\boldx) = \boldJ\boldnabla H$ is the symplectic part and $\boldf_d(\boldx)= -\boldR\boldnabla H$ is the dissipative part of the vector field. This is in agreement with a Helmholtz decomposition, where a smooth vector field is split into a divergence-free field and a gradient field \citep{Glotzl2023}.

\section{METHOD}\label{sec:3_method}
\vspace{-2mm}




\subsection{Odd kernel}

Consider a reproducing kernel which satisfies $k(\boldx,\boldz) = k(-\boldx,-\boldz)$. Then 
\begin{equation}\label{eq:odd_kernel}
    k_{\text{odd}}(\boldx,\boldz) = \frac{1}{2} (k(\boldx,\boldz) - k(\boldx,-\boldz)) \quad \in \R
\end{equation}
is an odd reproducing kernel with an associated RKHS \citep{Krejnik2012}.
Any function $f$ in the RKHS defined by $k_{\text{odd}}$ will then be odd, since $k_{\text{odd}}(-\boldx,\boldz) = -k_{\text{odd}}(\boldx,\boldz)$ and therefore $f(-\boldx) = -f(\boldx)$.

\subsection{Curl-free kernels and divergence-free kernels}

The curl-free reproducing kernel $\boldK_c(\boldx,\boldz) = \boldG_c(\boldx-\boldz) \in \R^{n\times n}$ and the divergence-free reproducing kernel $\boldK_d(\boldx,\boldz) = \boldG_d(\boldx-\boldz) \in \R^{n\times n}$ can be derived from a scalar shift-invariant reproducing kernel  ${k(\boldx, \boldz) = g(\boldx - \boldz) \in \R}$ \citep{Lowitzsch2002,Fuselier2006} by using
\begin{align}
    \boldG_c(\boldx) &= -\boldnabla \boldnabla^{\T}g(\boldx)\\
    \boldG_d(\boldx) &= (\boldnabla^2\eye_n -\boldnabla \boldnabla^{\T})g(\boldx)
\end{align}
where $\boldnabla^2 = \boldnabla^\T\boldnabla$. The curl-free kernel derived from the Gaussian kernel 
\begin{equation}\label{eq:scalar_gaussian_kernel}
    k_{\sigma}(\boldx,\boldz) = e^{-\frac{\| \boldx - \boldz \|^2}{2 \sigma^2}}\quad \in \R
\end{equation}
with kernel width ${\sigma > 0}$ is given by \citep{Fuselier2006}
\begin{equation}\label{eq:curl_free_kernel_from_gaussian}
    \boldG_{c}(\boldx) = -\boldnabla \boldnabla^{\T}g_\sigma(\boldx) = \frac{1}{\sigma^2} e^{-\frac{\boldx^{\T} \boldx}{2 \sigma^2}} \left( \eye_n - \frac{\boldx \boldx^{\T}}{\sigma^2} \right)
\end{equation}

The curl-free kernel can be written as
\begin{equation}\label{eq:curl_free_kernel_base}
    \boldG_{c}(\boldx) = -\boldnabla \boldnabla^{\T}g(\boldx) = \bb -\boldnabla \frac{\partial g(\boldx)}{\partial x_1} \; \dots \; -\boldnabla \frac{\partial g(\boldx)}{\partial x_n} \eb
\end{equation}
where it is seen that each column is a gradient of a scalar field ${\partial g(\boldx)/\partial x_i}$. In the case that $n=3$, the gradient of a scalar field is curl-free, which is the background for the term curl-free kernel. 

From ${\boldnabla^{\T}(\boldnabla^2\eye_n - \boldnabla \boldnabla^{\T}) = \boldnabla^{\T}\boldnabla^2 - \boldnabla^2\boldnabla^{\T} = 0}$ it follows that the divergence of $\boldG_d(\boldx)$ is zero. 

Define the scalar fields $\phi_i(\boldx) = {\partial g(\boldx)/\partial x_i}$. Then $\boldG_c = [\boldnabla \phi_1, \ldots, \boldnabla \phi_n]$. Any function in the RKHS of $\boldG_c$ will then be given by $\boldf_c(\boldx) = \sum_{i=1}^N \boldG_{c}(\boldx - \boldx_i)\bolda_i$, which gives
\begin{align}
    \boldf_c(\boldx) 
    = - \boldnabla \phi(\boldx)
\end{align}
where $\bolda_i = [a_{i1},\ldots,a_{in}]^\T$ and
\begin{equation}
    \phi(\boldx) = \sum_{i=1}^N\sum_{j=1}^N a_{ij} \phi_j(\boldx - \boldx_i)
\end{equation}
It is seen that the vector field $\boldf_c(\boldx)$ is the gradient of the scalar field $-\phi(\boldx)$. 


\subsection{Symplectic kernel as divergence-free kernel}

\citet{Boffi2022} proposed a symplectic kernel for adaptive prediction of Hamiltonian dynamics. The symplectic kernel $\boldK_{s}(\boldx,\boldz) = \boldG_{s}(\boldx - \boldz)$ is based on the curl-free kernel in \eqref{eq:curl_free_kernel_base}, and is given by 
\begin{equation}\label{eq:symplectic_kernel_base}
    \boldG_{s}(\boldx) 
    = \boldJ \boldG_{c}(\boldx) \boldJ^{\T}
    = - \boldJ \boldnabla \boldnabla^{\T}g(\boldx) \boldJ^{\T}
\end{equation}
The symplectic kernel is divergence-free, which follows from $\boldnabla^{\T} \boldJ \boldnabla = 0$ where it is used that $\boldJ$ is skew-symmetric. To verify that functions in the resulting RKHS are symplectic, it is used that a function in the RKHS of $\boldG_s$ is given by
\begin{equation}
\boldf(\boldx) = \sum_{i=1}^N  \boldG_{s}(\boldx - \boldx_i) \bolda_i = -\boldJ \boldnabla \sum_{i=1}^N  \boldnabla^{\T}g(\boldx - \boldx_i) \boldc_i
\end{equation}
where ${\boldc_{i} = \boldJ^{\T} \bolda_{i}}$. This results in the Hamiltonian dynamics ${\boldf(\boldx) = \boldJ \boldnabla H(\boldx)}$ where the Hamiltonian is the scalar function 
\begin{equation}\label{eq:learned_hamiltonian}
    {H}(\boldx) = -\sum_{i = 1}^{N} \boldnabla^{\T} g(\boldx - \boldx_{i}) \boldc_{i}
\end{equation}
Using the Gaussian kernel \eqref{eq:scalar_gaussian_kernel}, the symplectic kernel is \citep{Boffi2022}
\begin{equation}\label{eq:symplectic_kernel_from_gaussian}
    \boldG_{s}(\boldx) = \frac{1}{\sigma^2} e^{-\frac{\boldx^{\T} \boldx}{2 \sigma^2}} \boldJ \left( \eye_n - \frac{\boldx \boldx^{\T}}{\sigma^2} \right) \boldJ^{\T}
\end{equation}

\subsection{Odd curl-free and odd symplectic kernel}
The odd curl-free and the odd symplectic kernels were introduced by \cite{Smith2023}. The odd curl-free kernel is 
\begin{multline}\label{eq:symmetric_curl_free_kernel}
    \boldK_{c,o}(\boldx,\boldz) = \frac{1}{2\sigma^2} \biggl( e^{-\frac{\| \boldx - \boldz \|^2}{2\sigma^2}} \biggl( \eye_n - \frac{(\boldx - \boldz)(\boldx - \boldz)^{\T}}{\sigma^2} \biggr) \\ - e^{-\frac{\| \boldx + \boldz \|^2}{2\sigma^2}} \biggl( \eye_n - \frac{(\boldx + \boldz)(\boldx + \boldz)^{\T}}{\sigma^2} \biggr) \biggr)
\end{multline}
while the odd symplectic kernel is given by
\begin{equation}\label{eq:odd_symplectic_kernel}
    \boldK_{s,o}(\boldx,\boldz) = \frac{1}{2} \left( \boldG_{s}(\boldx-\boldz) - \boldG_{s}(\boldx+\boldz) \right)
\end{equation}
where ${\boldG_{s}}$ is defined in \eqref{eq:symplectic_kernel_from_gaussian}.



\subsection{RFF approximation of the odd curl-free and odd symplectic kernels}

\citet{Rahimi2008} showed how scalar shift-invariant kernels could be approximated using random Fourier features. This was extended to matrix kernels in \citep{Brault2016,Minh2016}. In \citep{Smith2024}, random Fourier features for the odd curl-free and odd symplectic kernels were derived. The matrix-valued odd curl-free kernel function can be approximated by the matrix-valued feature map ${\boldK_{c,o}(\boldx,\boldz) \approx \boldPsi_{c,o}(\boldx)^{\T} \boldPsi_{c,o}(\boldz)}$ where
\begin{equation}\label{eq:odd_curl_free_rff_map}
    \boldPsi_{c,o}(\boldx) = \frac{1}{\sqrt{d}} 
    \begin{bmatrix}
        \sin(\boldw_{1}^{\T}\boldx)\boldw_{1}^{\T}\\
        \vdots\\
        \sin(\boldw_{d}^{\T}\boldx)\boldw_{d}^{\T}
    \end{bmatrix}
    \in \R^{d \times n}
\end{equation}
The matrix-valued odd symplectic kernel function can be approximated by ${\boldK_{s,o}(\boldx,\boldz) \approx \boldPsi_{s,o}(\boldx)^{\T} \boldPsi_{s,o}(\boldz)}$ where
\begin{equation}\label{eq:odd_symplectic_rff_map}
    \boldPsi_{s,o}(\boldx) = \frac{1}{\sqrt{d}} 
    \begin{bmatrix}
        \sin(\boldw_{1}^{\T}\boldx)(\boldJ \boldw_{1})^{\T}\\
        \vdots\\
        \sin(\boldw_{d}^{\T}\boldx)(\boldJ \boldw_{d})^{\T}
    \end{bmatrix}
    \in \R^{d \times n}
\end{equation}
The weights ${\{ \boldw_i \}_{i=1}^{d} \in \Rn}$ are drawn i.i.d. with distribution ${\calN\hspace{-.8mm}\left(\zero,\sigma^{-2} \eye_n\right)}$.

\subsection{Solution of the Helmholtz decomposition}
The Helmholtz decomposition of the vector field into a symplectic part and a dissipative part as in \eqref{eq:Helmholtz decomposition} is achieved by separately modeling the symplectic and the divergence-free components using their corresponding random features. Then \eqref{eq:rkhf_f_using_rff}, \eqref{eq:odd_curl_free_rff_map}, and \eqref{eq:odd_symplectic_rff_map} are used to define
\begin{align}
    \boldf_{d}(\boldx) &= \boldPsi_{c,o}(\boldx)^{\T} \boldalpha\\
    \boldf_{s}(\boldx) &= \boldPsi_{s,o}(\boldx)^{\T} \boldbeta
\end{align}
The learned Helmholtz decomposition can be written as
\begin{equation}\label{eq:helmholtz_decomposition_rff}
    \boldf(\boldx) = \boldPsi_{c,o}(\boldx)^{\T} \boldalpha + \boldPsi_{s,o}(\boldx)^{\T} \boldbeta
\end{equation}
Given a dataset ${\calZ = \{ (\boldx_{i},\boldxdot_{i}) \in \Rn \times \Rn \}_{i=1}^{N}}$ the Helmholtz decomposition can be learned using the vector-valued regularized least-squares problem
\begin{equation}\label{eq:rff_helmholtz_least_squares}
    \begin{split}
        \min_{\boldalpha,\boldbeta \in \Rd} \frac{1}{N} \sum_{i=1}^{N} \| \boldPsi_{c,o}(\boldx_i)^{\T} \boldalpha + \boldPsi_{s,o}(\boldx_i)^{\T} \boldbeta - \boldxdot_{i} \|^{2}\\
        + \lambda_{1} \boldalpha^{\T} \boldalpha + \lambda_{2} \boldbeta^{\T} \boldbeta
    \end{split}
\end{equation}
where ${\lambda_1,\lambda_2 > 0}$ are the regularization parameters. This is solved using the following matrix formulation. Define the matrix
\begin{equation}
    \boldPhi = \bb \boldPsi_{c,o}(\boldx_1) & \dots & \boldPsi_{c,o}(\boldx_N)\\[4pt] \boldPsi_{s,o}(\boldx_1) & \dots & \boldPsi_{s,o}(\boldx_N) \eb \in \R^{2d \times nN}
\end{equation}
and the vectors ${\boldxi = \bb \boldalpha^{\T} & \boldbeta^{\T} \eb^{\T}}$ and ${\boldX = \bb \boldxdot_1^{\T} \dots \boldxdot_N^{\T} \eb^{\T}}$. Then
the optimization problem in \eqref{eq:rff_helmholtz_least_squares} can be written as
\begin{equation}
    \min_{\boldxi \in \R^{2d}} \frac{1}{N} \left( \boldPhi^{\T} \boldxi - \boldX \right)^{\T} \left( \boldPhi^{\T} \boldxi - \boldX \right) + \boldxi^{\T} \boldLambda \boldxi
\end{equation}
where the matrix ${\boldLambda \in \R^{2d \times 2d}}$ is a diagonal matrix containing the regularization parameters ${\lambda_{1}}$ and ${\lambda_{2}}$ corresponding to ${\boldalpha}$ and ${\boldbeta}$. The minimum is found when the gradient with respect to ${\boldxi}$ is zero
\begin{equation}
    \boldxi^* = \bb \boldalpha^*\\ \boldbeta^* \eb = \inv{\left( \boldPhi \boldPhi^{\T} + N \boldLambda \right)} \boldPhi \boldX
\end{equation}
and the function value of the optimal vector field is given by \eqref{eq:helmholtz_decomposition_rff}.
\section{EXPERIMENTS}\label{sec:4_experiments}
\vspace{-2mm}
Two simulation experiments were performed to validate the proposed method, referred to as the Helmholtz model, and to compare it to a Gaussian model, which uses a Gaussian separable kernel feature map \citep{Singh2021}. Each simulation experiment used data sets generated by the simulation of a dissipative Hamiltonian system. The data sets had a limited number of samples and included noise. The hyperparameter ${\sigma}$ of the Gaussian kernel and the regularization hyperparameter ${\lambda}$ of the optimization problem greatly influence the performance of the models, so they were tuned using a cross-validation method \citep{Kohavi1995,Smith2024}, solved using a genetic algorithm in MATLAB.

\subsection{Mass spring damper}
\begin{figure*}[t!]
    \centering
    \begin{subfigure}[h]{0.232\textwidth}
    \centering
        \includegraphics[width=\textwidth]{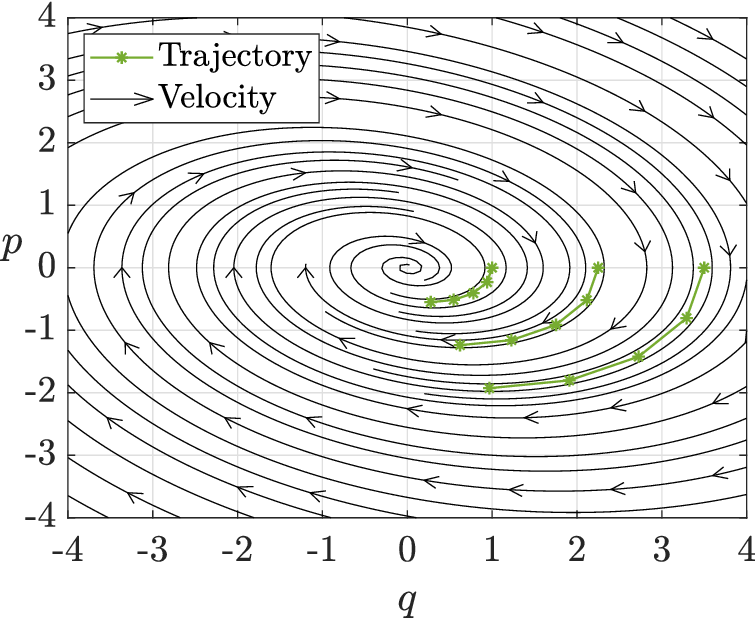}
        \caption{True system}
        \label{fig:harmonic_oscillator_true_model}
    \end{subfigure}
    \hfill
    \begin{subfigure}[h]{0.232\textwidth}
    \centering
        \includegraphics[width=\textwidth]{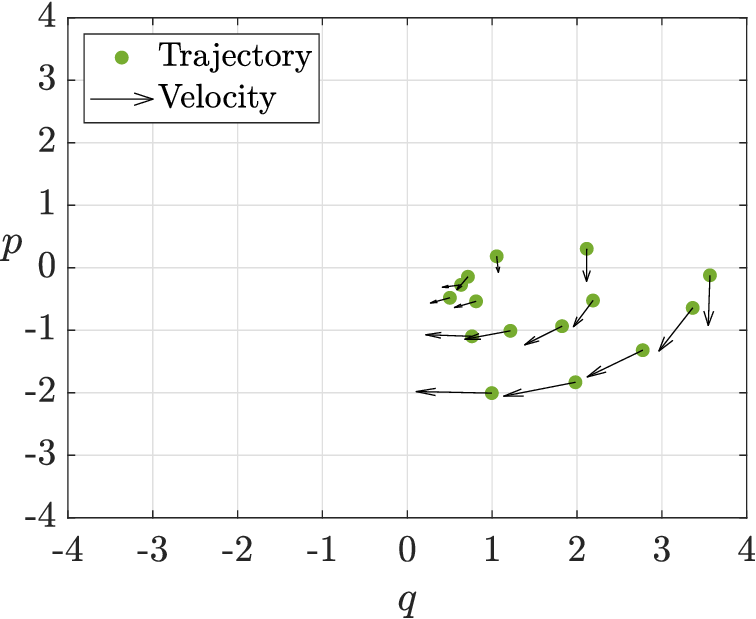}
        \caption{Data set}
        \label{fig:harmonic_oscillator_data_set}
    \end{subfigure}
    \hfill
    \begin{subfigure}[h]{0.232\textwidth}
    \centering
        \includegraphics[width=\textwidth]{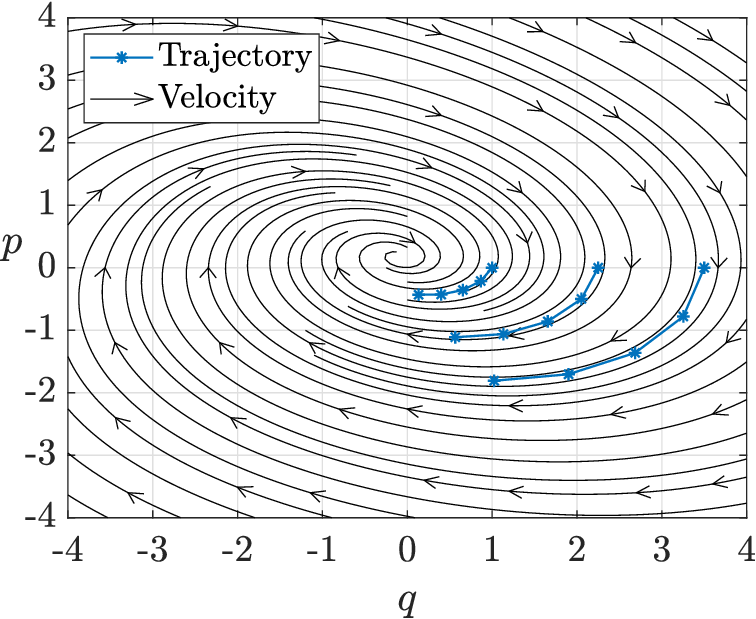}
        \caption{Learned Gaussian model}
        \label{fig:harmonic_oscillator_gaussian}
    \end{subfigure}
    \hfill
    \begin{subfigure}[h]{0.232\textwidth}
    \centering
        \includegraphics[width=\textwidth]{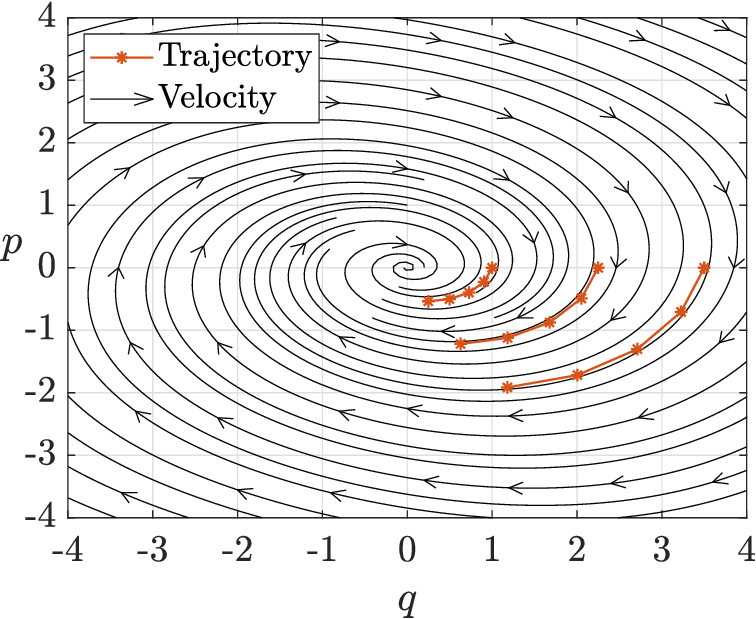}
        \caption{Learned Helmholtz model}
        \label{fig:harmonic_oscillator_helmholtz}
    \end{subfigure}
    \vspace{-1.5mm}
    \caption{Stream and trajectory plots for the mass-spring-damper and extracted data set, and the resulting learned models using the Gaussian separable kernel and the Helmholtz model.}
    \label{fig:harmonic_oscillator}
\end{figure*}

\vspace{-1mm}
The first simulation experiments used an undamped mass-spring system with mass ${m > 0}$ and spring constant ${k > 0}$. The position of the mass is ${x \in \R}$. The kinetic energy is ${T = \frac{1}{2} m \xdot^2}$ and the potential energy is ${U = \frac{1}{2} k x^2}$. The generalized coordinate is ${q = x}$, the generalized momentum is ${p = m \xdot}$, and the state vector is $\boldx = [q,p]^\T$. The Hamiltonian for the system is $H(q,p) = \frac{1}{2} \frac{p^2}{m} + \frac{1}{2} k q^2$. A dissipative force is given by $F_d = -d \qdot = -\frac{d}{m} p$. The resulting dynamics are then 
\begin{equation}
    \qdot = \frac{p}{m}, \qquad \pdot = -k q - \dfrac{d}{m} p
\end{equation}
Figure~\ref{fig:harmonic_oscillator_true_model} shows phase curves of the true system as described with parameters ${m = 0.5}$, ${k = 1}$ and ${d = 0.25}$. Figure~\ref{fig:harmonic_oscillator_true_model} also shows three trajectories that were generated by simulating the true system with the three initial conditions ${\boldx_{1,0} = \bb 1, 0\eb^{\T}}$, ${\boldx_{2,0} = \bb 2.25, 0\eb^{\T}}$, and ${\boldx_{3,0} = \bb 3.5, 0\eb^{\T}}$. The time step was ${h = \SI{0.25}{\second}}$ and the system was simulated for a short time ${t \in \bb 0, 1 \eb}$ seconds, which resulted in a limited dataset of ${5}$ data points for each trajectory, and in total ${N = 15}$ data points. The velocities ${\boldxdot}$ were sampled at each

\begin{figure*}[t!]
    \centering
    \begin{subfigure}[h]{0.232\textwidth}
    \centering
        \includegraphics[width=\textwidth]{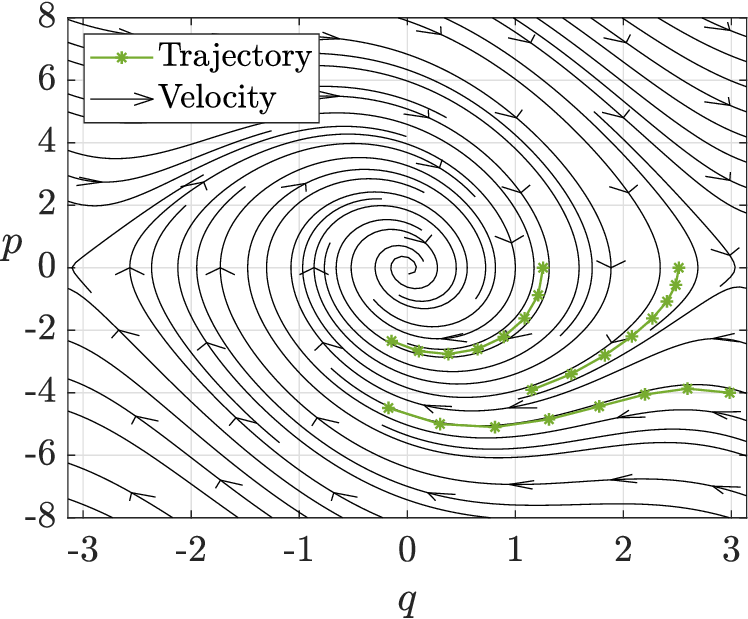}
        \caption{True system}
        \label{fig:simple_pendulum_true_model}
    \end{subfigure}
    \hfill
    \begin{subfigure}[h]{0.232\textwidth}
    \centering
        \includegraphics[width=\textwidth]{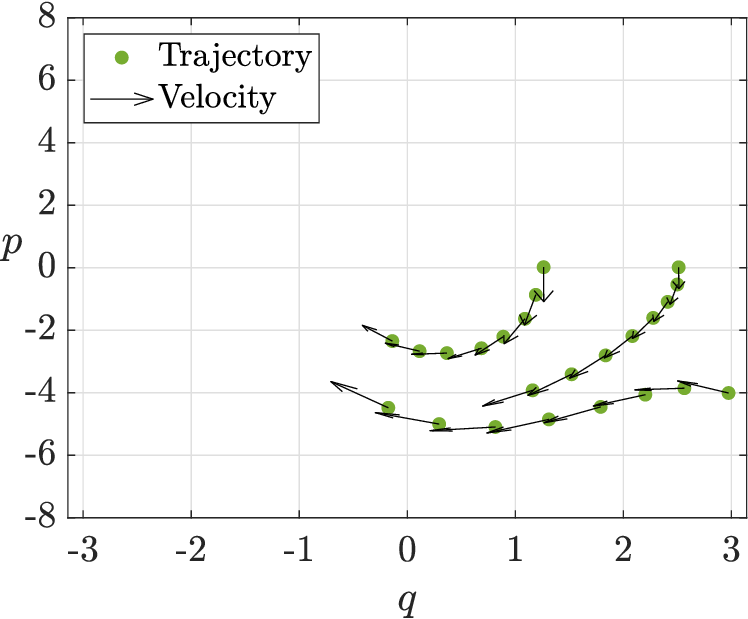}
        \caption{Data set}
        \label{fig:simple_pendulum_data_set}
    \end{subfigure}
    \hfill
    \begin{subfigure}[h]{0.232\textwidth}
    \centering
        \includegraphics[width=\textwidth]{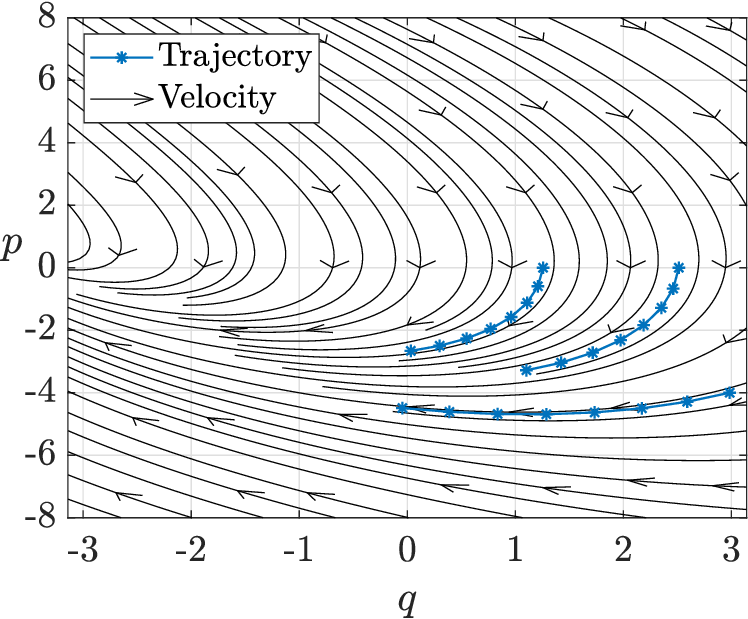}
        \caption{Learned Gaussian model}
        \label{fig:simple_pendulum_gaussian}
    \end{subfigure}
    \hfill
    \begin{subfigure}[h]{0.232\textwidth}
    \centering
        \includegraphics[width=\textwidth]{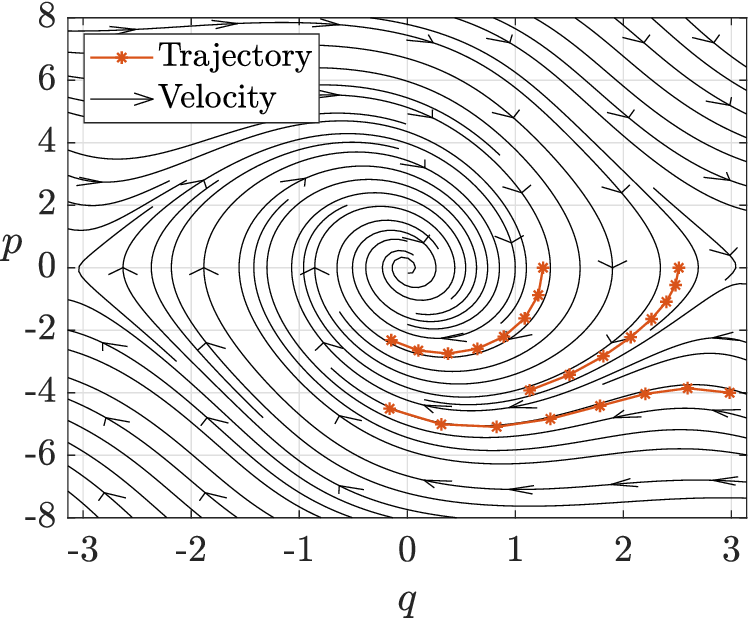}
        \caption{Learned Helmholtz model}
        \label{fig:simple_pendulum_helmholtz}
    \end{subfigure}
    \vspace{-1.5mm}
    \caption{Stream and trajectory plots for the damped pendulum and extracted data set, and the resulting learned models using the Gaussian separable kernel and the Helmholtz model.}
    \label{fig:simple_pendulum}
\end{figure*}

trajectory point, and zero mean Gaussian noise with standard deviation ${\sigma_n = 0.1}$ was added to the trajectory and velocity data. Figure~\ref{fig:harmonic_oscillator_data_set} shows the resulting data set.

\subsection{Damped pendulum}
\vspace{-1mm}
A simple pendulum with mass ${m}$ and length $l$ moves with the angle ${\theta}$. The generalized coordinate is ${q = \theta}$, the kinetic energy is ${T = \frac{1}{2} m l^2 \qdot^2}$ and the potential energy is ${U = m g l (1 - \cos(q))}$, where ${g}$ is the gravitational constant. The generalized momentum is ${p = \frac{\partial L}{\partial \qdot} = m l^2 \qdot}$ and the Hamiltonian is $H(q,p) = \frac{p^2}{2 m l^2} + m g l (1 - \cos(q))$. A dissipative force is given by $F_d =  -d \qdot = -\frac{d}{m l^2} p$. The dynamics for the damped pendulum are then 
\begin{equation}
    \qdot = \frac{p}{m l^2}, \qquad \pdot = - m g l \sin(q) -\dfrac{d}{m l^2} p
\end{equation}
Figure~\ref{fig:simple_pendulum_true_model} shows phase curves of the true system with parameters ${m = 1}$, ${l = 1}$, ${d = 1.2}$, and ${g = 9.81}$. Three trajectories were generated, as the system was simulated with the three initial conditions ${\boldx_{1,0} = \bb 2\pi / 5, 0\eb^{\T}}$, ${\boldx_{2,0} = \bb 4\pi / 5, 0\eb^{\T}}$, and ${\boldx_{3,0} = \bb 19\pi / 20, -4 \eb^{\T}}$. The time step was set to ${h = \SI{0.1}{\second}}$ and the system was simulated for a short time ${t \in \bb 0, 0.7 \eb}$ seconds, giving a limited dataset of ${8}$ data points for each trajectory, and ${N = 24}$ data points in total. The velocities ${\boldxdot}$ were sampled at each trajectory point, and zero mean Gaussian noise with standard deviation ${\sigma_n = 0.01}$ was added to the trajectory and velocity data. Figure~\ref{fig:simple_pendulum_data_set} shows the resulting data set.

\subsection{Numerical evaluation}
\vspace{-1mm}
The results showed that the Helmholtz model gave a significant improvement over the Gaussian model for the mass-spring-damper. This can be seen in Figure~\ref{fig:harmonic_oscillator_gaussian} and Figure~\ref{fig:harmonic_oscillator_helmholtz} where the Helmholtz model is centered about the origin and the Gaussian model is not. In the second simulation experiment, the Gaussian model failed to recreate the true vector field of the pendulum model due to overfitting, while the Helmholtz model recreated the true system. This can be seen in Figure~\ref{fig:simple_pendulum_gaussian} and Figure~\ref{fig:simple_pendulum_helmholtz}.

The learned models were evaluated by taking the mean squared error (MSE) over the training set and a separate test trajectory to test the generalized performance of the learned models. To generate the test trajectory, the mass-spring-damper system was simulated with the initial condition ${\boldx_0 = \bb 2, 0 \eb^{\T}}$ and the damped pendulum was simulated with the initial condition ${\boldx_0 = \bb \pi/2, 0 \eb^{\T}}$. Both systems were simulated for ${t \in \bb 0, 20 \eb}$ seconds to give a test set with a longer time horizon than the limited training sets. Table~\ref{tab:mse_results} shows the resulting MSE for the mass-spring-damper and the damped pendulum. The Gaussian model produces decent results for the training MSE on the mass-spring-damper, but for all other MSE values, there is a significant improvement when using the Helmholtz model.

\renewcommand{\captionwidth}{5cm}
\begin{table}[htb!]
    \caption{Numerical evaluation}
    \vspace{-1mm}
    \centering
    \begin{tabular}{@{\extracolsep\fill}lcccc}
        \toprule
        & \multicolumn{2}{c}{Mass spring damper} & \multicolumn{2}{c}{Damped pendulum}\\
        \cmidrule(lr){2-3} \cmidrule(lr){4-5}
        System & Training MSE\hspace{-2.5mm} & Test MSE & Training MSE\hspace{-2.5mm} & Test MSE\\
        \midrule
        Gaussian\hspace{-2mm} & 0.0496\hspace{-2.5mm} & 0.0921 & 0.1291\hspace{-2.5mm} & 17.725\\
        Helmholtz\hspace{-2mm}  & 0.0419\hspace{-2.5mm} & 0.0119 & 0.0007\hspace{-2.5mm} & 0.0003\\
        \bottomrule
    \end{tabular}
    \label{tab:mse_results}
\end{table}
\section{CONCLUSION}\label{sec:5_conclusion}
\vspace{-2mm}
This paper has demonstrated that dissipative Hamiltonian dynamical systems can be learned using functions in an RKHS with RFF. The proposed method uses a Helmholtz decomposition to create an additive model with a symplectic part learned via a symplectic kernel and a dissipative part via a curl-free kernel, both modified for odd symmetry to enhance performance. Validation through simulation experiments confirmed the method's ability to accurately capture and generalize the dynamics of mechanical systems, surpassing a baseline implementation. Future work may explore its application to more complex systems, control-oriented learning, and removing the need for generalized momenta data.
\vspace{-1mm}

\begin{ack}
    \vspace{-2mm}
    This work was partially funded by the Research Council of Norway under MAROFF Project Number 295138.
\end{ack}


\bibliography{refs}             

\begin{thebibliography}{28}
\providecommand{\natexlab}[1]{#1}
\providecommand{\url}[1]{\texttt{#1}}
\providecommand{\urlprefix}{URL }
\expandafter\ifx\csname urlstyle\endcsname\relax
  \providecommand{\doi}[1]{doi:\discretionary{}{}{}#1}\else
  \providecommand{\doi}{doi:\discretionary{}{}{}\begingroup \urlstyle{rm}\Url}\fi

\bibitem[{Ahmadi et~al.(2018)Ahmadi, Topcu, and Rowley}]{Ahmadi2018}
Ahmadi, M., Topcu, U., and Rowley, C. (2018).
\newblock {Control-Oriented Learning of Lagrangian and Hamiltonian Systems}.
\newblock In \emph{American Control Conference (ACC)}, 520--525.
\newblock \doi{10.23919/ACC.2018.8431726}.

\bibitem[{Arnold(1989)}]{Arnold1989}
Arnold, V.I. (1989).
\newblock \emph{{Mathematical Methods of Classical Mechanics}}.
\newblock Graduate Texts in Mathematics. Springer, New York, 2 edition.
\newblock \doi{10.1007/978-1-4757-2063-1}.

\bibitem[{Aronszajn(1950)}]{Aronszajn1950}
Aronszajn, N. (1950).
\newblock {Theory of Reproducing Kernels}.
\newblock \emph{Transactions of the American Mathematical Society}, 68(3), 337--404.

\bibitem[{Boffi et~al.(2022)Boffi, Tu, and Slotine}]{Boffi2022}
Boffi, N.M., Tu, S., and Slotine, J.J.E. (2022).
\newblock Nonparametric adaptive control and prediction: theory and randomized algorithms.
\newblock \emph{Journal of Machine Learning Research}, 23(281), 1--46.

\bibitem[{Brault et~al.(2016)Brault, Heinonen, and Buc}]{Brault2016}
Brault, R., Heinonen, M., and Buc, F. (2016).
\newblock {Random Fourier Features For Operator-Valued Kernels}.
\newblock In \emph{Proceedings of The 8th Asian Conference on Machine Learning}, volume~63 of \emph{Proceedings of Machine Learning Research}, 110--125. PMLR.

\bibitem[{Brunton and Kutz(2022)}]{Brunton2022}
Brunton, S.L. and Kutz, J.N. (2022).
\newblock \emph{Data-Driven Science and Engineering: Machine Learning, Dynamical Systems, and Control}.
\newblock Cambridge University Press, 2 edition.
\newblock \doi{10.1017/9781009089517}.

\bibitem[{Chen et~al.(2020)Chen, Zhang, Arjovsky, and Bottou}]{Chen2020}
Chen, Z., Zhang, J., Arjovsky, M., and Bottou, L. (2020).
\newblock {Symplectic Recurrent Neural Networks}.
\newblock In \emph{International Conference on Learning Representations}.

\bibitem[{Fuselier(2006)}]{Fuselier2006}
Fuselier, E.J. (2006).
\newblock {Refined Error Estimates for Matrix-Valued Radial Basis Functions}.
\newblock PhD thesis, Texas A\&M University.

\bibitem[{Glötzl and Richters(2023)}]{Glotzl2023}
Glötzl, E. and Richters, O. (2023).
\newblock Helmholtz decomposition and potential functions for $n$-dimensional analytic vector fields.
\newblock \emph{Journal of Mathematical Analysis and Applications}, 525(2), 127138.
\newblock \doi{10.1016/j.jmaa.2023.127138}.

\bibitem[{Goldstein et~al.(2002)Goldstein, Poole, and Safko}]{Goldstein2002}
Goldstein, H., Poole, C.P., and Safko, J.L. (2002).
\newblock \emph{{Classical Mechanics}}.
\newblock Pearson, 3 edition.

\bibitem[{Greydanus et~al.(2019)Greydanus, Dzamba, and Yosinski}]{Greydanus2019}
Greydanus, S., Dzamba, M., and Yosinski, J. (2019).
\newblock {Hamiltonian Neural Networks}.
\newblock In \emph{Proceedings of the Conference on Neural Information Processing Systems (NeurIPS)}.

\bibitem[{Hairer et~al.(2006)Hairer, Wanner, and Lubich}]{Hairer2006}
Hairer, E., Wanner, G., and Lubich, C. (2006).
\newblock \emph{{Geometric Numerical Integration: Structure-Preserving Algorithms for Ordinary Differential Equations}}.
\newblock Springer Series in Computational Mathematics. Springer Berlin, Heidelberg, 2 edition.
\newblock \doi{10.1007/3-540-30666-8}.

\bibitem[{Kohavi(1995)}]{Kohavi1995}
Kohavi, R. (1995).
\newblock {A Study of Cross-Validation and Bootstrap for Accuracy Estimation and Model Selection}.
\newblock In \emph{Proceedings of the 14th International Joint Conference on Artificial Intelligence, {IJCAI-95}}, 1137--1143.

\bibitem[{Krejnik and Tyutin(2012)}]{Krejnik2012}
Krejnik, M. and Tyutin, A. (2012).
\newblock {Reproducing Kernel Hilbert Spaces With Odd Kernels in Price Prediction}.
\newblock \emph{IEEE Transactions on Neural Networks and Learning Systems}, 23(10), 1564--1573.
\newblock \doi{10.1109/TNNLS.2012.2207739}.

\bibitem[{Lowitzsch(2002)}]{Lowitzsch2002}
Lowitzsch, S. (2002).
\newblock \emph{Approximation and Interpolation Employing Divergence-free Radial Basis Functions with Applications}.
\newblock Ph.D. thesis, {Texas A\&M University}.

\bibitem[{Micchelli and Pontil(2005)}]{Micchelli2005}
Micchelli, C.A. and Pontil, M. (2005).
\newblock {On Learning Vector-Valued Functions}.
\newblock \emph{Neural Computation}, 17(1), 177--204.
\newblock \doi{10.1162/0899766052530802}.

\bibitem[{Minh(2016)}]{Minh2016}
Minh, H.Q. (2016).
\newblock {Operator-Valued Bochner Theorem, Fourier Feature Maps for Operator-Valued Kernels, and Vector-Valued Learning}.
\newblock \doi{10.48550/arXiv.1608.05639}.
\newblock ArXiv preprint arXiv:1608.05639 [cs.LG].

\bibitem[{Minh and Sindhwani(2011)}]{Minh2011}
Minh, H.Q. and Sindhwani, V. (2011).
\newblock {Vector-valued Manifold Regularization}.
\newblock In \emph{Proceedings of the 28th International Conference on Machine Learning (ICML1 1)}, 57--64. Omnipress.
\newblock \doi{10.5555/3298023.3298180}.

\bibitem[{Ortega et~al.(2008)Ortega, van~der Schaft, Castaños, and Astolfi}]{Ortega2008}
Ortega, R., van~der Schaft, A., Castaños, F., and Astolfi, A. (2008).
\newblock Control by interconnection and standard passivity-based control of port-{H}amiltonian systems.
\newblock \emph{IEEE Transactions on Automatic Control}, 53(11), 2527--2542.

\bibitem[{Rahimi and Recht(2008)}]{Rahimi2008}
Rahimi, A. and Recht, B. (2008).
\newblock {Uniform Approximation of Functions with Random Bases}.
\newblock In \emph{2008 46th Annual Allerton Conference on Communication, Control, and Computing}, 555--561.
\newblock \doi{10.1109/ALLERTON.2008.4797607}.

\bibitem[{Schölkopf et~al.(2001)Schölkopf, Herbrich, and Smola}]{Scholkopf2001}
Schölkopf, B., Herbrich, R., and Smola, A.J. (2001).
\newblock {A Generalized Representer Theorem}.
\newblock In \emph{Computational Learning Theory}, 416--426. Springer Berlin Heidelberg.
\newblock \doi{10.1007/3-540-44581-1\_27}.

\bibitem[{Sindhwani et~al.(2018)Sindhwani, Tu, and Khansari}]{Sindhwani2018}
Sindhwani, V., Tu, S., and Khansari, M. (2018).
\newblock {Learning Contracting Vector Fields For Stable Imitation Learning}.
\newblock \doi{10.48550/arXiv.1804.04878}.
\newblock ArXiv preprint arXiv:1804.04878 [cs.RO].

\bibitem[{Singh et~al.(2021)Singh, Richards, Sindhwani, Slotine, and Pavone}]{Singh2021}
Singh, S., Richards, S.M., Sindhwani, V., Slotine, J.J.E., and Pavone, M. (2021).
\newblock {Learning stabilizable nonlinear dynamics with contraction-based regularization}.
\newblock \emph{The International Journal of Robotics Research}, 40(10-11), 1123--1150.
\newblock \doi{10.1177/0278364920949931}.

\bibitem[{Smith and Egeland(2023)}]{Smith2023}
Smith, T. and Egeland, O. (2023).
\newblock {Learning of Hamiltonian Dynamics with Reproducing Kernel Hilbert Spaces}.
\newblock \doi{10.48550/arXiv.2312.09734}.
\newblock ArXiv preprint arXiv:2312.09734 [cs.RO].

\bibitem[{Smith and Egeland(2024)}]{Smith2024}
Smith, T. and Egeland, O. (2024).
\newblock {Learning Hamiltonian Dynamics with Reproducing Kernel Hilbert Spaces and Random Features}.
\newblock \doi{10.48550/arXiv.2404.07703}.
\newblock ArXiv preprint arXiv:2404.07703 [cs.LG].

\bibitem[{Sosanya and Greydanus(2022)}]{Sosanya2022}
Sosanya, A. and Greydanus, S. (2022).
\newblock {Dissipative Hamiltonian Neural Networks: Learning Dissipative and Conservative Dynamics Separately}.
\newblock \doi{10.48550/arXiv.2201.10085}.
\newblock ArXiv preprint arXiv:2201.10085 [cs.LG].

\bibitem[{Zhong et~al.(2020{\natexlab{a}})Zhong, Dey, and Chakraborty}]{Zhong2020dissipative}
Zhong, Y.D., Dey, B., and Chakraborty, A. (2020{\natexlab{a}}).
\newblock {Dissipative SymODEN: Encoding Hamiltonian Dynamics with Dissipation and Control into Deep Learning}.
\newblock \doi{10.48550/arXiv.2002.08860}.
\newblock ArXiv preprint arXiv:2002.08860 [cs.LG].

\bibitem[{Zhong et~al.(2020{\natexlab{b}})Zhong, Dey, and Chakraborty}]{Zhong2020}
Zhong, Y.D., Dey, B., and Chakraborty, A. (2020{\natexlab{b}}).
\newblock {Symplectic ODE-Net: Learning Hamiltonian Dynamics with Control}.
\newblock In \emph{Proceedings of the International Conference on Learning Representations (ICLR)}.

\end{thebibliography}


\end{document}